\documentclass[runningheads]{llncs}
\usepackage{graphicx}
\usepackage{amssymb}
\usepackage{subfigure}
\usepackage{cite}
\usepackage{booktabs}
\usepackage{bm}
\usepackage[misc,geometry]{ifsym}
\begin{document}
%
%\title{Contribution Title\thanks{Supported by organization x.}}
%%
%%\titlerunning{Abbreviated paper title}
%% If the paper title is too long for the running head, you can set
%% an abbreviated paper title here
%%
%\author{First Author\inst{1}\orcidID{0000-1111-2222-3333} \and
%Second Author\inst{2,3}\orcidID{1111-2222-3333-4444} \and
%Third Author\inst{3}\orcidID{2222--3333-4444-5555}}
%%
%\authorrunning{F. Author et al.}
%% First names are abbreviated in the running head.
%% If there are more than two authors, 'et al.' is used.
%%
%\institute{Princeton University, Princeton NJ 08544, USA \and
%Springer Heidelberg, Tiergartenstr. 17, 69121 Heidelberg, Germany
%\email{lncs@springer.com}\\
%\url{http://www.springer.com/gp/computer-science/lncs} \and
%ABC Institute, Rupert-Karls-University Heidelberg, Heidelberg, Germany\\
%\email{\{abc,lncs\}@uni-heidelberg.de}}
%%
%\maketitle              % typeset the header of the contribution
%%
\title{Revisiting 3D Context Modeling with Supervised Pre-training for Universal Lesion Detection in CT Slices}

\author{Shu Zhang\inst{1},
Jincheng Xu\inst{1}\thanks{This work was done when Jincheng Xu was an intern at Deepwise AI Lab.},
Yu-Chun Chen\inst{2},
Jiechao Ma\inst{2},
Zihao Li\inst{2},
Yizhou Wang\inst{1,3,4}
\and Yizhou Yu\inst{2,5}$^{\left(\textrm{\Letter}\right)}$
}
%index{Zhang, Shu} 
%index{Xu, Jincheng} 
%index{Chen, Yu-Chun} 
%index{Ma, Jiechao} 
%index{Li, Zihao} 
%index{Wang, Yizhou} 
%index{Yu, Yizhou}

\institute{
{Department of  Computer Science, Peking University, Beijing, China}
\and {Deepwise AI Lab, Beijing, China\\ \email{yizhouy@acm.org}}
\and{Advanced Institute of Information Technology, Peking University, Hangzhou, China}
\and{Center on Frontiers of Computing Studies, Peking University, Beijing, China}
\and{The University of Hong Kong, Pokfulam, Hong Kong}
}
\maketitle

\begin{abstract}
%The abstract should briefly summarize the contents of the paper in
%15--250 words.

Universal lesion detection from computed tomography (CT) slices is important for comprehensive disease screening. 
Since each lesion can locate in multiple adjacent slices, 3D context modeling is of great significance for developing automated lesion detection algorithms. 
In this work, we propose a \textit{Modified Pseudo-3D Feature Pyramid Network} (MP3D FPN) that leverages depthwise separable convolutional filters and a group transform module (GTM) to efficiently extract 3D context enhanced 2D features for universal lesion detection in CT slices.
To facilitate faster convergence, a novel 3D network pre-training method is derived using solely large-scale 2D object detection dataset in the natural image domain. 
We demonstrate that with the novel pre-training method, the proposed MP3D FPN achieves state-of-the-art detection performance on the DeepLesion dataset (\textbf{3.48\%} absolute improvement in the sensitivity of FPs@0.5), significantly surpassing the baseline method by up to \textbf{6.06\%} (in MAP@0.5) which adopts 2D convolution for 3D context modeling. 
Moreover, the proposed 3D pre-trained weights can potentially be used to boost the performance of other 3D medical image analysis tasks.

\keywords{Lesion Detection  \and 3D Context Modeling \and 3D Network Pre-training.}
\end{abstract}
\section{Introduction}

With its high-resolution image and low cost, CT scan is critical in clinical decision and holds the key for making precise medical-care accessible to everyone around the world. 
%How to improve its detecting ability for universal lesions is salient for effective use of medical resources. 
Recently, deep learning methods have been introduced  to detect lesions in CT slices~\cite{3DCE,MSB, Retina, MVP, MULAN}. 
Since it is difficult to distinguish lesions within a single axial slice, exploiting sufficient 3D context for accurate detection in volumetric CT data has emerged as a significant research focus.

Various architectures have been proposed for proper modeling of 3D context from neighboring CT slices. 
Yan \textit{et al.}~\cite{3DCE} adopts a late fusion strategy which stacked 2D features of neighboring slices to build 3D context enhanced features. 
Although the pseudo-3D contextual information has provided prominent performance gain~\cite{3DCE,MSB, Retina, MVP, MULAN}, its late fusion strategy leads to notable losses of context information from early stages of the network.  
A direct way to address these issues is to employ 3D convolutions which introduce inter-slice connections hierarchically to learn 3D representations end to end. 3D convolutional filters can well preserve the 3D structure and texture information,  
but intensive memory and computation demands hinder its wide application in the universal lesion detection problem. 
What's worse, although 3D network pre-training has raised significant research attention~\cite{MG, Med3D,P3D,ACS}, the lack of good pre-trained 3D models makes it even harder to achieve good performance with 3D based detectors.

In this paper, we focus on the problem of universal lesion detection in CT slices, where multiple adjacent CT slices are taken into consideration to localize 2D lesions for the target slice.
We aim to develop a generic and efficient 3D backbone for 2D lesion detection with enhanced context modeling ability from multiple CT slices and devise a supervised pre-training method to boost its performance. 
Specifically, pseudo-3D convolutional filters~\cite{P3D} which use depth-wise separable convolution are adopted to reduce the memory and computation overhead. 
The backbone in our method is a Modified Pseudo-3D ResNet (MP3D ResNet), which extracts context enhanced 3D features from multiple neighboring CT slices (9 in our case) and then converted the 3D features into 2D ones with a group transform module (GTM) for further 2D lesion detection in the target slice. 
Then, we feed backbone features extracted from MP3D ResNet into the neck of Feature Pyramid Network (FPN) to form the MP3D FPN for effective multi-scale detection. 
Finally, to facilitate efficient training of the MP3D FPN, we designed a novel supervised pre-training method, which exploits supervised signals from large-scale 2D natural image object detection dataset to pre-train the proposed MP3D detector. In summary, the main contributions of our paper are three folds:

\paragraph{1.} We have proposed a generic framework to employ 3D network for 2D lesion detection in CT slices. The proposed MP3D FPN is computational and memory efficient, and it achieves state-of-the-art performance on the DeepLesion dataset.
% without any auxiliary supervision.  %can be readily used in other 3D medical image analysis problems.
\paragraph{2.} We have derived a novel and effective way to adopt 2D natural images to pre-train 3D network with supervised labels, whose pre-trained weights can potentially benefit other 3D medical image analysis tasks (e.g. segmentation).
\paragraph{3.} We have conducted comprehensive experiments to explore the effects of pre-trained weights for deep medical image analysis. The results suggest that pre-trained weights can not only lead to faster convergence in all sized datasets, but also help to achieve better results in smaller-scale ones.

\section{Methodology}

\begin{figure}[!tbp]
\centering
\includegraphics[width=\textwidth]{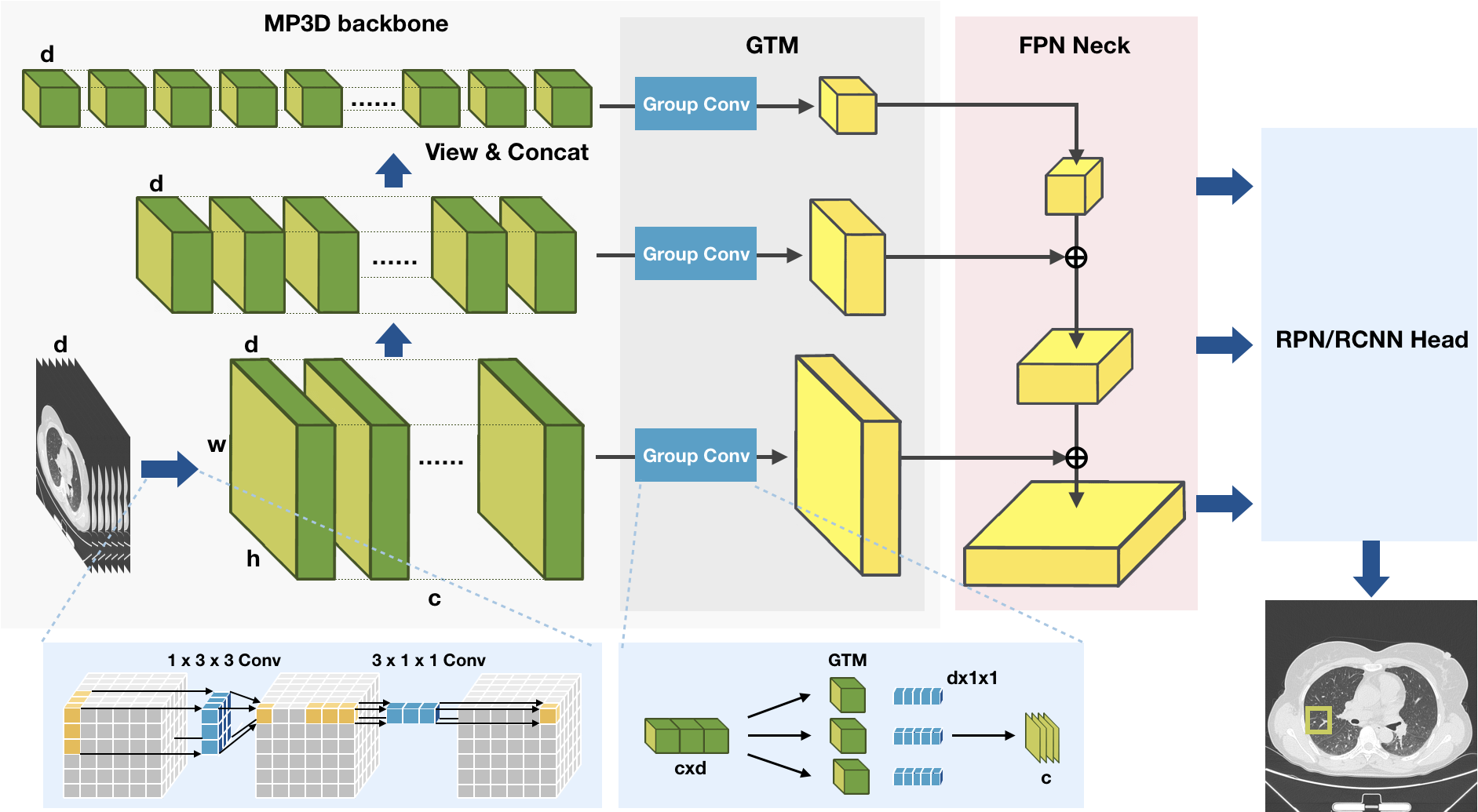}
\caption{Overview of the proposed MP3D FPN. MP3D ResNet extracts context enhanced 3D features and converts them to 2D ones with a group transform module (GTM). These context enhanced 2D features are then fed into the FPN neck and the RPN/RCNN head for further 2D lesion detection. The MP3D FPN is pre-trained on Microsoft COCO object detection dataset~\cite{COCO}. } \label{flowchart}
\end{figure}

Fig.\ref{flowchart} gives an overview of the proposed lesion detection framework.
The proposed MP3D FPN comprises an MP3D ResNet as the backbone, a 2D FPN~\cite{FPN} as the neck and a 2D RPN/RCNN head.
%To facilitate 2D bounding box detection in CT slices, both the neck and head are set the same as general 2D FPN networks. 
The MP3D ResNet takes multiple consecutive CT slices (e.g. 9) as input and generates 3D feature maps which bear the ability of 3D context modeling. 
Then a conversion block (GTM) further transforms the 3D feature maps into 2D ones for further 2D detection. 
Detailed architecture designs of the proposed MP3D backbone and the novel supervised pre-training scheme will be elaborated in the following sections.

\subsection{3D Context Modeling with an MP3D ResNet Backbone}

In this work, we explore to employ 3D convolutions for effective 3D context modeling in the problem of lesion detection from consecutive CT slices (e.g. 9 slices). To advance the time and memory efficiency of normal 3D ResNet, we adopt the Pseudo-3D Residual Network (P3D ResNet)~\cite{P3D} as the prototype of our backbone network.
The pseudo-3D convolution simulates $3\times 3\times 3$ convolution with $1\times 3\times 3$ filter on axial-view slices plus $3\times 1\times 1$ filter to build inter-slice connections on adjacent CT slices.

Lesion detection in CT slices aims to predict 2D bounding boxes in a certain slice, thus it requires 2D feature maps corresponding to the target slice for further prediction. 
Therefore, we need to convert the 3D feature maps to 2D ones for further prediction, meanwhile preserving the precise information of the target CT slice for accurate localization and classification. 
The designed Modified Pseudo-3D Residual Network (MP3D ResNet) highlights two aspects of modifications to fulfill such demands:
1) Instead of conducting isotropic pooling as in the original P3D ResNet, we neglect pooling operation in the inter-slice dimension. 
2) A group transform module is introduced to generate the desired 2D feature maps from the context enhanced 3D features.

Neglecting pooling operation in the inter-slice dimension can help to preserve precise information of the target slice. In the meantime, since the number of input slices (e.g. 9) is rather small, we can get enough receptive field in the inter-slice dimension without downsampling. 
Regarding 2D feature map conversions, Fang \textit{et al.}~\cite{PFN} proposed to extract $C$ 2D feature maps ($1\times 1\times H\times W$) corresponding to the center slice and concatenate them to form the converted 2D feature map of size ($C\times H\times W$).
However, this method can not fully exploit the 3D context information resided in other adjacent slices.

We, on the other hand, propose a group transform module (GTM) instead to includes all slice's features to compensate for the information loss.
Specifically, we view 3D features ($C\times D\times H\times W$) into 2D ($CD\times H\times W$) and apply a group convolutional layer with the group size of $C$ (every $D$ channel is a group) to fuse all neighboring features to yield the final 2D feature maps ($C\times H\times W$). 

\begin{figure}[!tbp]
\includegraphics[width=\textwidth]{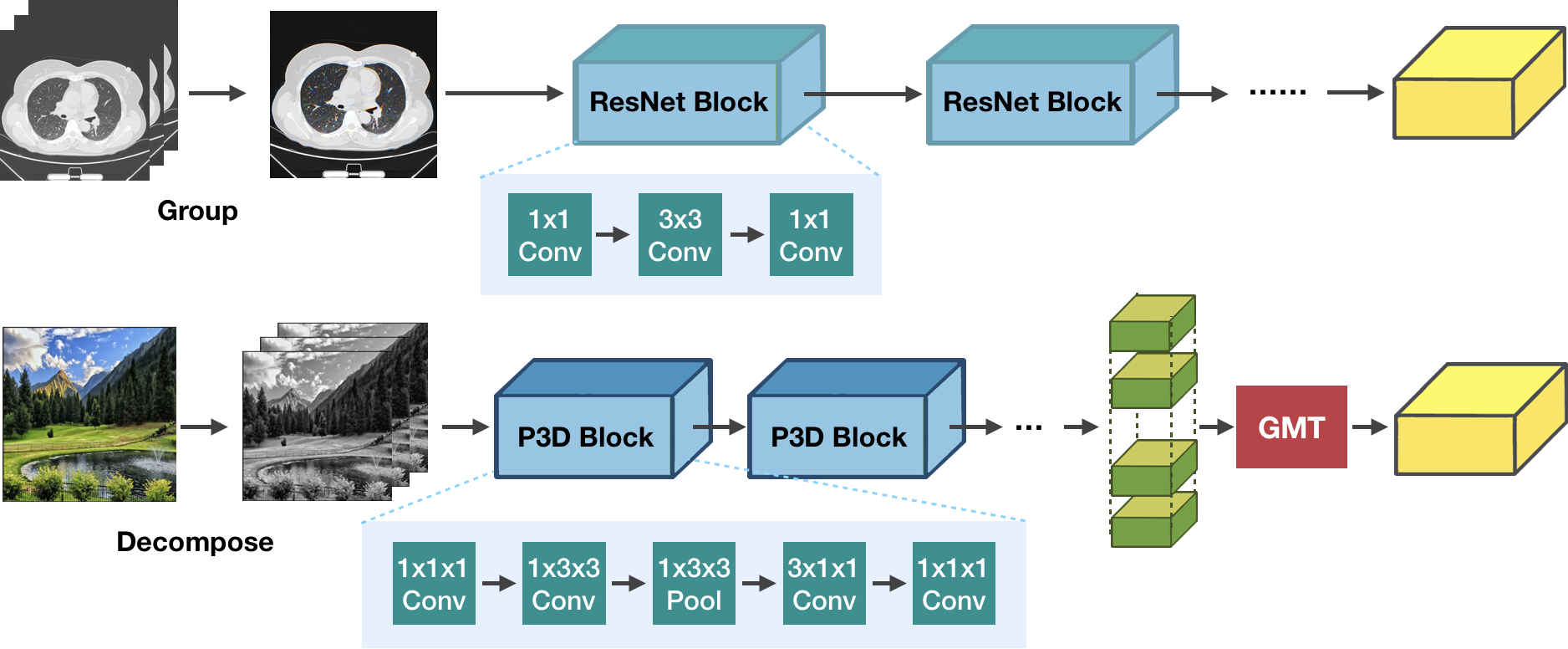}
\caption{Comparison between 1) using 2D Image-Net pre-trained weights for multi-slice medical image analysis and 2) decomposing 2D natural image to simulate multi-slice medical image for 3D network pre-training. } \label{pretrain}
\end{figure}

\subsection{Supervised 3D Pre-training with COCO Dataset}

\begin{table}[!htbp]
    \caption{Sensitivities (\%) at various FPs per image on the test set of DeepLesion. $\mathbf{^*}$ indicates re-implementation of 3DCE using ResNet-50 FPN with the same configuration as our MP3D FPN.}
    \label{tab1}
    \centering
    \footnotesize % fontsize
    \setlength{\tabcolsep}{8pt}% column separation
    \renewcommand{\arraystretch}{1.2}%row space 
        \centering
        \label{tab:sample_1}
        \begin{tabular}{l|ccccccccc}
            \hline
            \textbf{Methods} & $\mathbf{0.5}$ & $\mathbf{1}$ & $\mathbf{2}$ & $\mathbf{4}$ & $\mathbf{MAP@0.5}$ \\
            %\cline{2-9}% partial hline from column i to column j
            \hline
            \hline
            3DCE, 27 slices\cite{3DCE} & 52.86 & 64.80 & 74.84 & 84.38 &-\\
            MSB, 3 slices\cite{MSB} & 67.00 & 76.80 & 83.70 & 89.00 &-\\
            RetinaNet, 3 slices\cite{Retina} & 72.15 & 80.07 & 86.40 & 90.77 &-\\
	    MVP-Net, 9 slices\cite{MVP} & 73.83 & 81.82 & 87.60 & 91.30 &-\\		
	    MULAN, 9 slices\cite{MULAN} & \textbf{76.12} & \textbf{83.69} & \textbf{88.76} & \textbf{92.30} &-\\
            \hline
            FPN+3DCE, 3 slices$\mathbf{^*}$ &68.52 & 77.59 & 83.91  & 88.33 & 64.41\\
            FPN+3DCE, 9 slices$\mathbf{^*}$  &74.06 & 82.00 & 87.58 & 91.56 & 70.28\\
            FPN+3DCE, 27 slices$\mathbf{^*}$  &\textbf{74.67} & \textbf{82.89} & \textbf{88.17} & \textbf{91.62} & \textbf{70.82}\\
            \hline
            \textbf{MR3D FPN, 9 slices} & 79.09 & 84.84 & 89.18 & 92.06  & 76.57\\
            \textbf{MP3D FPN, 9 slices} & \textbf{79.60} & \textbf{85.29} & \textbf{89.61}  & \textbf{92.45}  & \textbf{76.87} \\
            \hline
            \textbf{Imp over} MULAN, 9 slices & $\uparrow$\textbf{3.48}  & $\uparrow$1.60 & $\uparrow$0.85 & $\uparrow$0.15&- \\
 
        \end{tabular}
\end{table}

Supervised pre-training from natural images has proven to be an effective way for 2D medical image transfer learning\cite{3DCE,MSB,Retina,MVP,MULAN, Chest}. This indicates that using supervised pre-training models from another domain can actually benefit the medical image analysis application. 
What's more, compared to self-supervised signals, we believe that supervised labels which carry the semantic information could enable the model to learn semantically invariant and discriminative features more effectively.
Therefore, in this section, we aim to develop a method to exploit supervised labels from large-scale 2D natural image object detection dataset (e.g. coco~\cite{COCO}) to pre-train our MP3D FPN.

Previous works~\cite{3DCE} have shown that by grouping 3 consecutive CT slices (which is natively 3D data) as a 3-channel RGB image, we can boost the detection performance with Image-Net pre-trained weights, indicating the feasibility of simulating RGB natural image with natively 3D CT slices.
This inspires us to reversely decompose the 3 channels of natural RGB images into 3 consecutive CT slices, and train an MP3D FPN with such simulated 3D data. 
Fig~\ref{pretrain} illustrates a comparison of the two correlative strategies. 
For implementation details, we train the MP3D FPN on COCO dataset for 72 epochs and the final weights are used to initialize MP3D ResNet. 
To drive the network to learn useful 3D contextual features from inter-slice connections, it is essential to keep the resolution in the inter-slice dimension unchanged for all stages of the backbone.% for the MP3D ResNet backbone when pre-training on COCO. 
The MP3D detector trained with a slice number of 3 can be used to initialize lesion detectors which takes variable number of slices as network input.

\section{Experiments}

\subsection{Experimental Setup}
\noindent\textbf{Dataset and Metric:} The NIH DeepLesion is a large-scale dataset for lesion detection, which contains 32,735 lesions on 32,120 axial CT slices captured from 4,427 patients. DeepLesion is splitted into training (70\%), validation (15\%), and test (15\%) sets. We evaluate our MP3D FPN and all the compared methods on the test set by reporting the mean average precision (MAP@0.5) and average sensitivities at different false positives (FPs) per image.

\noindent\textbf{Implementation Details:} As in~\cite{Retina}, the Hounsfield units (HU) are clipped into the range of $[-1024,1050]$. 
We interpolate in the z-axis to normalize the intervals of all CT slices to 2.5mm.
Anchor scales are set to $\{16, 32, 64, 128, 256\}$ in FPN.
Apart from horizontal and vertical flip, we resize the image to different scales of $\{448, 512, 576\}$ for data augmentation.
MP3D-63 with group normalization\cite{GN} is used as the backbone in all our experiments, which has similar depth with the ResNet3D-50 model. The MP3D-63 model is derived from the conventional P3D-63\cite{P3D} model with the proposed modifications.  
Unless otherwise specified, the MP3D FPN takes 9 consecutive slices as input. 
We train all the models for 24 epochs at the base learning rate of 0.02, and reduce it by a factor of 10 after the 16-th and 22-th epoch (corresponding to the 2x learning schedule\cite{Rethinking} on COCO dataset). 
We conduct experiments on the NVIDIA TITAN V GPU with 12GB of memory, and mixed-precision training strategy is used in all our experiments to save memory.

\subsection{Comparison with State-of-the-arts}

Table \ref{tab1} presents the comparisons with the previous state-of-the-art (SOTA) methods.
Our model surpasses all the SOTA methods on sensitivities at different FPs and MAP@0.5, which includes 3DCE\cite{3DCE}, MSB\cite{MSB}, RetinaNet\cite{Retina}, MVP-Net\cite{MVP} and MULAN\cite{MULAN}.

Without using any auxiliary supervision, MP3D FPN outperforms MULAN, the previous SOTA which additionally employs multi task learning and a deeper backbone (DenseNet-121) to improve the detection accuracy, by up to \textbf{3.48\%} on the sensitivity of FPs@0.5.
We re-implement 3DCE with ResNet-50 FPN using the same configuration as our MP3D FPN for fair comparison. Our proposed MP3D achieve a performance gain of \textbf{6.05\%} on MAP@0.5 compared with this 2D convolution based context encoding method, demonstrating the superior 3D context modeling ability of our MP3D backbones. 
As shown in Table \ref{tab1}, MP3D FPN (248.93 GFLOPS, 45.16 $M$ Params) and MR3D FPN (Modified ResNet 3D, 415.81 GFLOPS, 64.03 $M$ Params) based detector achieve comparable results, but the MP3D based detector consumes much less time and memory.
This strongly proves the efficacy and the thrift of our MP3D model.

\subsection{Ablation Study}
We perform a number of ablations to probe into our MP3D FPN. The results are shown as follows:

\begin{table}[!t]
    \caption{Detection performance and computational cost with variable numbers of input slice. GFLOPS is used to characterize the computational cost.}
    \label{tab:slice_num}
    \centering
    \footnotesize % fontsize
    \setlength{\tabcolsep}{8pt}% column separation
    \renewcommand{\arraystretch}{1.2}%row space 
        \centering
        %\label{tab:slice_num}
        \begin{tabular}{lccccccccc}
            \textbf{Methods} & $\mathbf{0.5}$ & $\mathbf{1}$ & $\mathbf{2}$ & $\mathbf{4}$ & $\mathbf{MAP@0.5}$ & $\mathbf{GFLOPS}$\\
            \hline
            \hline
            MP3D, 5 slices  &76.86	&83.44&88.13 &91.54&75.01 & 156.84\\
	     MP3D, 7 slices &78.22	&84.45&88.90 &91.50&76.69 & 202.88\\
            MP3D, 9 slices &79.60 &85.29 & \textbf{89.61}  &92.45  &76.87 & 248.93\\
	      MP3D, 11 slices&\textbf{80.05}&\textbf{85.77}&89.55 &\textbf{92.45}&\textbf{77.64} & 294.97\\
        \end{tabular}
\end{table}

\noindent\textbf{Input Slices:} Table \ref{tab:slice_num} shows the performance of the MP3D detector when applying 5,7,9 and 11 slices as input. 
The detector achieves higher detection accuracy as more slices are used, meanwhile consuming more time and memory. MP3D with 7 slices as input get the best trade-off between effectiveness and efficiency. 

\begin{table}[!t]
    \caption{Comparison of different conversion modules and different pooling strategies for pre-training.}
    \label{tab:gtm}
    \centering
    \footnotesize % fontsize
    \setlength{\tabcolsep}{8pt}% column separation
    \renewcommand{\arraystretch}{1.2}%row space 
        \centering
        %\label{tab:gtm}
        \begin{tabular}{lccccccccc}
            \textbf{Methods} & $\mathbf{0.5}$ & $\mathbf{1}$ & $\mathbf{2}$ &  $\mathbf{4}$ & $\mathbf{MAP@0.5}$\\
            %\cline{2-9}% partial hline from column i to column j
            \hline
            \hline
            MP3D w/ CTM &79.18	&84.90 &88.96	  &91.90  &76.30 \\
            MP3D w/ GTM &\textbf{79.60} &\textbf{85.29} & \textbf{89.61}  & \textbf{92.45}  &\textbf{76.87}  \\
            \hline
            MP3D w/ isotropic pooling  &78.24	&84.41	   &88.82	&91.98  &75.06\\
            MP3D w/ proposed pooling &\textbf{79.60} &\textbf{85.29} & \textbf{89.61}  & \textbf{92.45}  &\textbf{76.87}  \\
        \end{tabular}
\end{table}

\noindent\textbf{Conversion Type:} Table \ref{tab:gtm} demonstrates the comparisons of proposed GTM with the center-cropping transform module (CTM), which is proposed by Fang \textit{et al.}~\cite{PFN}. The proposed GTM brings better results as it can efficiently aggregate information from all adjacent slices for further detection.

\subsection{Effectiveness of the 3D Pre-trained Model}

We conducted three groups of experiments to explore effectiveness of the pre-training method.

\noindent\textbf{Comparison to Isotropic Pooling:} In this work, to achieve 3D context modeling ability in the z-axis, we neglect pooling operation in the inter-slice dimension when pre-training the MP3D model on the Microsoft COCO dataset. We compared our proposed method to isotropic pooling for validation.  

The pre-trained model takes three slices as input. When training with isotropic pooling, the z-axis degenerates to a single slice after the first two pooling layers, preventing further 3D convolution layers from learning useful 3D contextual information. As shown in Table\ref{tab:gtm}, pre-trained weights learned from isotropic pooling gives worse results than the proposed method. This also proves that using decomposed natural image as input can actually helps the 3D model to gain context-encoding ability. Thus the learned weights can potentially be used to boost the performance of other 3D medical image analysis tasks.

\begin{table}[!t]
    \caption{Comparison of model performance with and without pre-training with different learning schedules. 1x, 2x and 6x indicates max training epochs of 12, 24 and 72 separately.}
    \label{tab:scratch}
    \centering
    \footnotesize % fontsize
    \setlength{\tabcolsep}{8pt}% column separation
    \renewcommand{\arraystretch}{1.2}%row space 
        \centering
        \label{tab:scratch}
        \begin{tabular}{lccccccccc}
            \textbf{Methods} & $\mathbf{0.5}$ & $\mathbf{1}$ & $\mathbf{2}$ & $\mathbf{4}$ & $\mathbf{MAP@0.5}$ \\
            %\cline{2-9}% partial hline from column i to column j
            \hline
            \hline
            MP3D  1x w/o pretrain   &	70.12&	78.00&	83.95&	88.23&67.60 \\
            MP3D  2x w/o pretrain  &	76.11&	82.65&	87.70&	91.17&74.00 \\
	    MP3D  6x w/o pretrain &\textbf{79.60} &	85.29&	89.26&	92.19& 76.75\\
	    \hline
           MP3D  1x w/ pretrain &78.02&	84.33&	88.84&	91.74&75.78\\
	    MP3D  2x w/ pretrain&79.58&        \textbf{85.29}&	\textbf{89.61}&	\textbf{92.45}&\textbf{76.87}\\
        \end{tabular}
\end{table}

\begin{table}[!t]
    \caption{Training with variable dataset sizes ($100\%$ to $20\%$). For simplicity, we present the results of MAP@0.5.} %All the 3DCE models use Image-Net pre-trained weights.}
    \label{tab4}
    \centering
    \footnotesize % fontsize
    \setlength{\tabcolsep}{8pt}% column separation
    \renewcommand{\arraystretch}{1.2}%row space 
        \centering
        \label{tab:datasize}
        \begin{tabular}{lcccccccc}  
            \textbf{Methods} & $\mathbf{100\%}$ & $\mathbf{80\%}$ & $\mathbf{60\%}$ & $\mathbf{40\%}$ & $\mathbf{20\%}$ \\
            %\cline{2-9}% partial hline from column i to column j
            \hline
            \hline
            3DCE  9 slices 2x    &70.28 &69.22 &67.08 &63.61 &57.02 \\
            3DCE  27 slices 2x  &70.82 &69.96 &68.08 &65.36 &58.82 \\
            \hline
            MP3D w/o pre-train 2x       &74.00 &71.58 &68.79 &63.40 &50.67 \\ 
            MP3D w/o pre-train 6x       &76.75 &75.43 &72.87 &68.14 &58.98 \\     
            MP3D w/ pre-train 2x & \textbf{76.87}  &\textbf{75.66}  &\textbf{73.33} &\textbf{71.07}  &\textbf{65.55} \\
     
        \end{tabular}
\end{table}

\noindent\textbf{Comparison to Training From Scratch:} 
He \textit{et al.}\cite{Rethinking} demonstrated that with sufficient training data (around 35k from its experiment) and longer training schedule (6x), models trained from scratch could achieve comparable results to models training with pre-trained weights. Therefore, we examined the effectiveness of our proposed pre-training method by comparing MP3D with pre-training to model trained from scratch with longer schedule.

As shown in Table\ref{tab:scratch}, when both trained for 1x learning schedule (12 epochs), MP3D with pre-trained weights significantly outperforms the one without pre-training, demonstrating faster convergence speed. And it turns out that with 2x learning schedule (24 epochs), model trained with the proposed pre-training weights can achieve comparable results with MP3D model trained from scratch with 6x learning schedule (72 epochs). These results validate the effectiveness of our proposed pre-training scheme.

\noindent\textbf{Performance on Variable Dataset Sizes:} In medical image analysis tasks, annotated data is often scarce. Therefore, it is appealing to gain a better understanding of the effects of pre-trained weights when dataset size is small. In this subsection, we compare the model performance of 2x, 6x training from scratch and 2x with pre-training on variable dataset sizes by randomly choosing 20\%, 40\%, 60\% and 80\% of the whole training data. Pre-training based models achieve better performance with less training time on all the cases, and the smaller the size of the dataset, the larger the gap. A dramatic drop of performance starts when training with only 40\% of the whole data. And when training with only 20\% of the dataset, which is around 4,500 images, the model trained with our proposed pre-trained weights achieves an absolute performance gain of 6.57\% on MAP@0.5, accounting for an 11\% relative gain.

\section{Conclusions}

In this paper, we propose a generic model architecture to exploit 3D network for 2D lesion detection in CT slices. 
The proposed MP3D FPN can reduce computation and memory cost while providing enhanced 3D context modeling ability. 
A simple yet effective way for 3D network pre-training is also derived to facilitate efficient training. 
Without sophisticated structures and multi-supervision signals, it significantly improves the detection performance on the DeepLesion dataset, surpassing all the SOTAs. 
We have proved the benefits of pre-trained weights for variable dataset size, and we expect that the MP3D ResNet along with its pre-trained weights can serve as a benchmark backbone for 3D medical image analysis, making contributions towards accessible precise medication.

\subsubsection{Acknowledgements.} This work is funded by National Key Research and Development Program of China (No. 2019YFC0118101), MOST-2018AAA0102004 and NSFC-61625201. We would like to thank Yemin Shi for valuable discussions.

\end{document}